%% file: main.tex
\documentclass[reprint,aps,pre]{revtex4-1}

\usepackage{amsmath,amsfonts,amscd,amssymb,graphicx,subeqnar}
\usepackage[bbgreekl]{mathbbol}

\input{z_vardef}
\usepackage{dcolumn}% Align table columns on decimal point
\usepackage{overpic}
\usepackage{color}
\usepackage{adjustbox}
\usepackage{float}
\usepackage{algorithm}
\usepackage{algpseudocode}
\algnewcommand\algorithmicinput{\textbf{Input:}}
\algnewcommand\Input{\item[\algorithmicinput]}

\usepackage{nomencl} % If needed
\usepackage{etoolbox}
\makenomenclature

\DeclareSymbolFontAlphabet{\mathbb}{AMSb}
\DeclareSymbolFontAlphabet{\mathbbl}{bbold}

\renewcommand{\nomgroup}[1]{%
  \ifstrequal{#1}{L}{\item[\textbf{Latin Symbols}]}{%
  \ifstrequal{#1}{G}{\item[\textbf{Greek Letters}]}{%
  \ifstrequal{#1}{A}{\item[\textbf{Acronyms}]}{}}}}

\renewcommand{\vec}[1]{\mathbf{#1}}

\newcommand{\norma}[1]{\left|\left|{#1}\right|\right|}

\usepackage{hyperref}

\makeatletter
\@ifundefined{sf@counterlist}{\def\sf@counterlist{}}{}
\makeatother

\usepackage[dvipsnames]{xcolor}

\definecolor{bluePolimi}{RGB}{22, 44, 80}
\definecolor{lightBluePolimi}{RGB}{91, 122, 172}
\definecolor{redPolimi}{RGB}{180, 0, 0}
\definecolor{greenPolimi}{RGB}{78, 172, 91}
\definecolor{green2}{RGB}{0, 110, 0}

\makeatletter
\hypersetup{colorlinks=true}
\AtBeginDocument{\@ifpackageloaded{hyperref}
  {\def\@linkcolor{blue}
   \def\@anchorcolor{red}
   \def\@citecolor{red}
   \def\@filecolor{red}
   \def\@urlcolor{redPolimi}
   \def\@menucolor{red}
   \def\@pagecolor{cyan}
\begingroup
  \@makeother\`%
  \@makeother\=%
  \edef\x{%
    \edef\noexpand\x{%
      \endgroup
      \noexpand\toks@{%
        \catcode 96=\noexpand\the\catcode`\noexpand\`\relax
        \catcode 61=\noexpand\the\catcode`\noexpand\=\relax
      }%
    }%
    \noexpand\x
  }%
\x
\@makeother\`
\@makeother\=
}{}}
\makeatother

\usepackage{comment}
\usepackage{tabularx}
\usepackage{booktabs} 

\begin{document}

\title{From Models To Experiments: Shallow Recurrent Decoder Networks on the DYNASTY Experimental Facility}

\author{Stefano Riva$^{a, b, *}$, Andrea Missaglia$^{c, b}$, Carolina Introini$^{b}$, J. Nathan Kutz$^{a}$, Antonio Cammi$^{b,d}$}
\affiliation{$^a$Autodesk Research, 6 Agar Street, London UK}
\email{stefano.riva@autodesk.com}

\affiliation{$^{b}$Politecnico di Milano, Deptartment of Energy, CeSNEF - Nuclear Engineering Division, 20156 Milan, Italy}
\affiliation{$^{c}$Texas A\&M University, Department of Nuclear Engineering, 423 Spence St, College Station, 77843-3133, TX, USA}
\affiliation{$^{d}$Emirates Nuclear Technology Center (ENTC), Department of Mechanical and Nuclear Engineering, Khalifa University, Abu Dhabi, 127788, United Arab Emirates}

\begin{abstract} 
    Shallow Recurrent Decoder networks are a novel scientific machine learning architecture, designed for state estimation from sparse measurements. It features important advantages compared to other architectures, including: the ability to use only few sensors for reconstructing the entire dynamics of a physical system; being agnostic to the positions of sensors; the ability to train on compressed data spanned by a reduced basis; the ability to measure a single field variable (easy to measure) and reconstruct coupled spatio-temporal fields that are unobservable; and minimal hyper-parameter tuning. This approach has already been adopted to a variety of problems, showing outstanding capabilities in providing accurate and reliable reduced models. However, its application (and subsequent validation) to a real experimental facility, using actual measured data collected on the physical system, is yet to be investigated. This work aims to fill this gap by applying the Shallow Recurrent Decoder architecture to DYNASTY, an experimental natural circulation loop designed to study this phenomenon for advanced nuclear reactors. The mathematical model for generating high-fidelity data is a validated RELAP5 model, while temperature measurements from the facility itself are used as input for the state estimation, and the mass flow rate is indirectly reconstructed. The key result of this work is that the architecture can reconstruct the unknown thermal-hydraulic state, even starting from only three temperature thermocouples, with high accuracy (the average relative percentage error remains below 1.5\%). Furthermore, the architecture can also predict unseen time instants beyond the temporal range of the available data, within the experimental uncertainty of $2.5$ K. These results prove the capability of the architecture to generalize in quasi-real-time, both in time and across different input conditions, a key capability required by digital twins.
\end{abstract}

\maketitle

% --------------------------------------
\section{Introduction}\label{sec:intro}
% --------------------------------------

Fast, accurate, and reliable \textit{digital twin} of nuclear and nuclear-related engineering plants \cite{Grieves_DT} are a topic of growing interest in the nuclear community \cite{10967494, en14144235, MENGYAN2024110491, mohanty_development_2021}, as they offer an innovative way of monitoring and controlling a nuclear power plant. Despite the absence of a unique definition, there is strong agreement that one of the key capabilities of models comprising a digital twin should be to infer the high-dimensional state of the system from sparse sensor measurements \cite{argaud_sensor_2018, cheng_efficient_2024, gong_efficient_2022}. Furthermore, such models must be accurate and computationally easy to solve, so that they can be utilized in online monitoring, control, or sensitivity analysis scenarios. Thus, an important property desired in digital twins is the ability to handle data coming from the physical system and combine their information with the background knowledge of mathematical models to, on one hand, account for un-modelled physics in the latter, and on the other hand, account for unobservable phenomena in the former \cite{haik_real-time_2023, riva2024multiphysics}.

A fundamental building block in the development of digital twins is state estimation, which aims to reconstruct the full multi-dimensional state of the system from a limited number of measurements, typically one-dimensional. This task is crucial for digital twins and is essentially an inverse problem, related to generating accurate surrogate models of the systems. In fact, tackling the state estimation problem at the high-fidelity level is computationally prohibitive \cite{quarteroni2015reduced}. To overcome this issue, over the years, many dimensionality reduction techniques have been developed, and the significant developments in data-driven science have further advanced research in this field \cite{zhang2025artificialintelligencereactorphysics}. In the literature, non-intrusive Reduced Order Modelling (ROM) \cite{rozza_model_2020}, such as the Gappy Proper Orthogonal Decomposition \cite{everson_karhunenloeve_1995,KarnikAbdo2024_Sensors} or the Generalised Empirical Interpolation Method \cite{ICAPP_plus2023, maday_generalized_2015} or a combination of the Dynamic Mode Decomposition with Kalman Filtering \cite{FALCONER2023133741}, can be considered a state-of-the-art approach to state estimation. These methods have the strong advantage of being generally interpretable and based on a solid mathematical background, with an \textit{a-priori} error estimation theory \cite{maday_parameterized-background_2014}. However, they are not well-suited to deal with strong nonlinear dynamics. Furthermore, they require \textit{ad hoc} extensions to handle the estimation of unobservable fields, as in \cite{ICAPP_plus2023,gong_parameter_2023,INTROINI2023109538}. This aspect is crucial for the application of these methods to real engineering systems, where measurements are typically limited to a few variables, whereas state estimation should provide a full-field spatial reconstruction of all relevant physical quantities.

In light of the limitations of traditional non-intrusive techniques and the rapid advancements in data science, machine learning (ML) techniques have also been proposed \cite{brunton_data-driven_2022}. In particular, for state estimation problems, a wide range of methods are available, including regression techniques such as decision trees \cite{ZHAO2025103527} and neural network-based techniques such as DeepONet \cite{Hossain:2025aa} and convolutional neural networks \cite{gong_reactor_2024, LeiteCNN2025}. Although these methods are very powerful in dealing with non-linear dynamics, they generally lack interpretability and require a significant amount of data for training. A more recent approach consists of combining linear dimensionality reduction methods like the Singular Value Decomposition (SVD) \cite{brunton_data-driven_2022} with ML architectures to build data-driven reduced models on a low-dimensional latent space rather than on the original high-dimensional one. Examples in the literature illustrate this idea: the reduced model can be built using Gaussian process regression \cite{HU2025121170}, autoencoders \cite{gong_data-enabled_2022}, or DeepONets \cite{Nair_Goza_2020}.

Focusing on this latter approach for time-dependent problems, recurrent neural networks \cite{hochreiter1997long} have been used successfully to build reduced models for forecasting. Among the available architectures, this work investigates the SHallow REcurrent Decoder (SHRED) network \cite{ebers2023leveraging, kutz_shallow_2024, williams2022data}. This architecture has been proven to be an accurate proxy to surrogate models of high-fidelity data, at a comparatively negligible computational cost. The SHRED architecture is an example of a data-driven methodology that builds a surrogate model through a training and learning process; mathematically, it is a generalization of the separation of variables method for solving Partial Differential Equations (PDEs) \cite{shredrom}. This technique has already been used to reconstruct high-dimensional data in the field of plasma dynamics \cite{kutz_shallow_2024}, innovative reactor concepts \cite{riva2024robuststateestimationpartial, riva2025efficientparametricstateestimation}, fluid dynamics \cite{moen2025mappingsurfaceheightdynamics, shredrom, CELIK2026124166}, and biological systems \cite{Rude2025}, showing an outstanding capability in generating accurate, reliable, and efficient surrogate models. More in detail, SHRED is a machine learning model that exploits the separation of time and parameters and space to build a reduced-order, non-linear model \cite{gao2025sparseidentificationnonlineardynamics}. The first separation is performed through the SVD \cite{lassila_model_2014, rozza_model_2020}, generating a suitable coordinate interpretable system \cite{brunton_data-driven_2022} in the form of a set of basis functions. Integrating the SVD into SHRED has been proven to be highly useful, enabling a compressive training approach and heavily reducing the computational costs of the learning process \cite{kutz_shallow_2024}. 

The authors have already analysed the SHRED architecture applied to a real system, a TRIGA Mark II reactor \cite{riva2025constrainedsensingreliablestate}, showing that SHRED can handle not only simulation data but also be deployed and tested on real data coming from the system itself. The referenced work dealt with a single scenario; it is yet to be studied how SHRED performs in an experimental facility, considering different parametric transient scenarios. Compared to synthetic data, experimental measures are not perfect; additionally, there are existing discrepancies between data and model \cite{maday_parameterized-background_2014}. To further explore the capabilities of SHRED applied to real-world physical systems, the DYNASTY (DYnamics of NAtural circulation for molten SalT internallY heated) experimental facility, built at Politecnico di Milano \cite{benzoni2023preliminary}, is now used as a test case. This system has been built to study the phenomenology of natural circulation dynamics for internally heated fluids, a topic of primary concern for the development of circulating fuel reactors such as the Molten Salt Fast Reactor \cite{GenIV-RoadMap}. In particular, the RELAP5 (R5) code \cite{fletcher1992relap5} is used to generate the high-dimensional data considering four different heating configurations and multiple parametric inputs. The R5 model then provides the mathematical background, and the SHRED architecture will be verified and validated against experimental data, which are the input for the architecture; focus will be on reconstruction for unknown parameters and temporal prediction outside the training dataset.

The paper will be structured as follows: Section \ref{sec: shred} focuses on a brief description of the SHRED architecture, then in Section \ref{sec: dynasty} the DYNASTY facility will be presented along with the associated R5 model; the numerical results are going to be discussed in Section \ref{sec: num-res} and the main conclusions will be drawn in Section \ref{sec: concl}.

% ----------------------------------------
\section{Shallow Recurrent Decoder Network}\label{sec: shred}
% ---------------------------------------
    
\begin{figure*}
    \centering
    \includegraphics[width=1\linewidth]{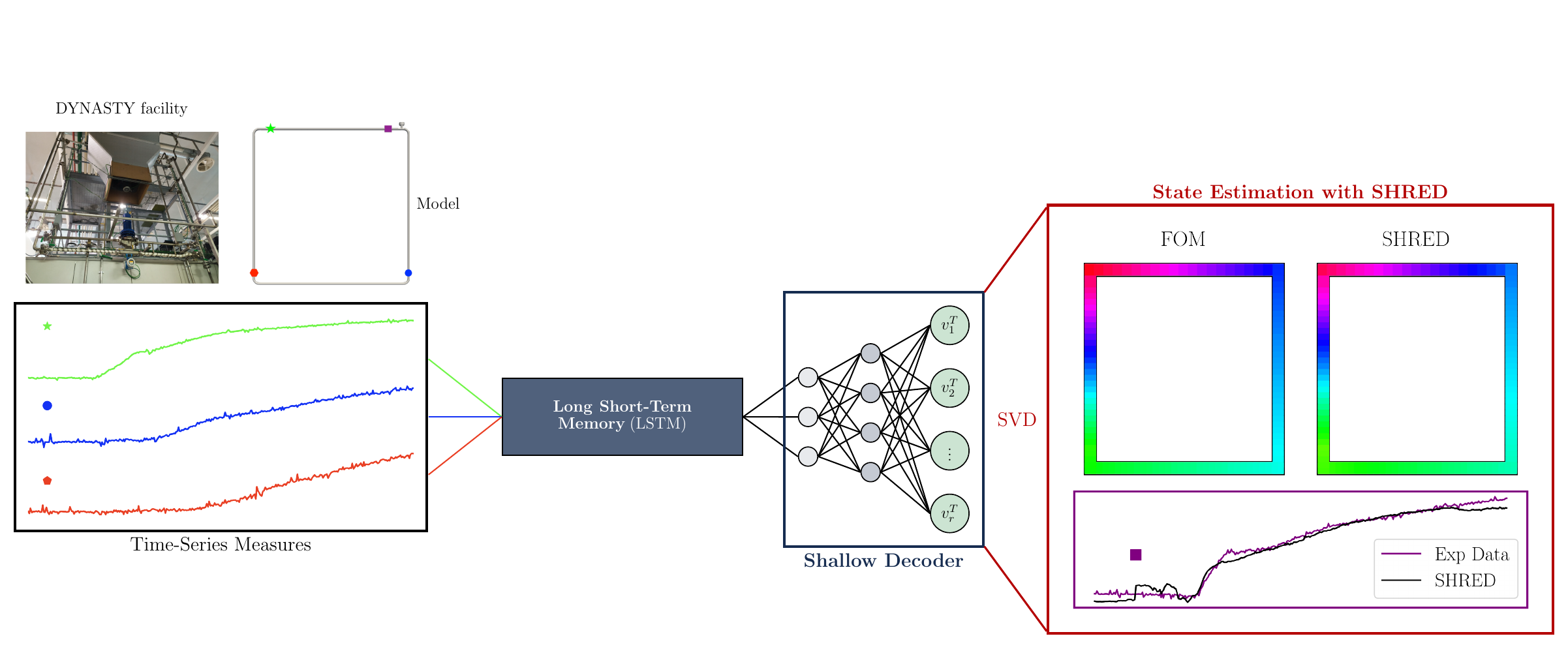}
    \caption{SHRED architecture applied to the DYNASTY facility. Three out of the four available thermocouples are used to measure the temperature in the fluid $T$, at fixed locations. The sensors time series are used to construct a latent temporal sequence model which is mapped to the compressive representations of all spatio-temporal field variables. The compressive representations can then be mapped to the original state space by the Singular Value Decomposition (SVD).}
    \label{fig: shred}
\end{figure*}

Shallow Recurrent Decoders are a class of neural network architectures specifically designed for state estimation, in particular for mapping sparse measurement trajectories $\vec{y}$ to the full state $\vec{u}$ \cite{ebers2023leveraging,kutz_shallow_2024,williams2022data}. There are two key components in SHRED: a Long Short-Term Memory (LSTM) network \cite{hochreiter1997long} and a Shallow Decoder Network (SDN) \cite{erichson2020shallow}, which can also be combined with a compressive training approach based on the Singular Value Decomposition (SVD) \cite{brunton_data-driven_2022}. The LSTM allows to learn the temporal dynamics contained in the sensor measurements and maps them to a latent space $\vec{z}$; then, the SDN decodes the latent representation back to the full state space, either directly or through a compressed representation $\vec{v}$ of the full state $\vec{u}$ obtained via SVD. This first reduction step is crucial to drastically reduce the computational costs of the training process \cite{kutz_shallow_2024}: the idea is to compress the high-dimensional data into a low-rank representation that captures the dominant spatial features of the system, by identifying a set of basis functions $\mathbb{U}$ through the SVD. The architecture is illustrated in Figure \ref{fig: shred}: the LSTM leverages Takens embedding theory \cite{takens1981lnm} to capture temporal patterns, while the SDN reconstructs latent representations that are subsequently decompressed using the SVD.

Compared to other data-driven state estimation methods like Gappy POD \cite{everson_karhunenloeve_1995}, Generalised Empirical Interpolation Method (GEIM) \cite{maday_generalized_2015} or other ML architectures \cite{gong_efficient_2022, LU2022114778, mohanty_development_2021}, SHRED offers several advantages: (1) it requires as few as three sensors, even randomly placed; (2) it enables training on compressed data, which drastically reduces computational costs and allows for laptop-level training in minutes \cite{kutz_shallow_2024}; (3) it effectively handles multi-physics (coupled) datasets even when only measurements from a single variable are available \cite{riva2024robuststateestimationpartial}; (4) it requires minimal hyperparameter tuning, as the same architecture can be applied across a wide range of problems \cite{kutz_shallow_2024,riva2024robuststateestimationpartial, shredrom,williams2022data}. Mathematically, SHRED can be viewed as a generalization of the classical separation of variables method for solving PDEs \cite{shredrom}, where the temporal/parametric and spatial components are separated and modeled using neural networks. The map is then expressed as follows:
\begin{subequations}
    \begin{align}
        \text{Compress. training: } \vec{u}(\vec{x}, t; \boldsymbol{\mu}) &\approx \mathbb{U} \cdot f_D\left(f_R\left(\vec{y}_L(t; \boldsymbol{\mu})\right)\right) \nonumber \\
        & \approx \mathbb{U} \cdot \hat{\vec{v}}(t; \boldsymbol{\mu}) \\
        \text{Full training: } \vec{u}(\vec{x}, t; \boldsymbol{\mu}) &\approx f_D\left(f_R\left(\vec{y}_L(t; \boldsymbol{\mu})\right)\right) \nonumber \\
        &\approx \hat{\vec{u}}(t; \boldsymbol{\mu})
    \end{align}
\end{subequations}
where $f_R$ is the LSTM network mapping the history of the $L$ measurements $\vec{Y}(t; \boldsymbol{\mu})=[\vec{y}(t), \vec{y}(t-\Delta t), \dots, \vec{y}(t-(L-1)\Delta t)]\in\mathbb{R}^{L\times s}$ to the latent representation $\vec{z}(t; \boldsymbol{\mu})\in\mathbb{R}^d$, $f_D$ is the SDN mapping $\vec{z}(t; \boldsymbol{\mu})$ to either the compressive representation $\hat{\vec{v}}(t; \boldsymbol{\mu})\in\mathbb{R}^r$ or directly to the full state $\hat{\vec{u}}(t; \boldsymbol{\mu})\in\mathbb{R}^{\mathcal{N}_h}$ directly, $s$ is the number of sensors, $\mathcal{N}_h$ is the number of spatial degrees of freedom, $\boldsymbol{\mu}$ is the set of parameters characterizing the system, and $r$ is the dimension of the compressed space (with $r << \mathcal{N}_h$). The learning process involves minimizing the mean squared error between the predicted and true states over the training dataset.

The SHRED architecture has been implemented in Python, using the PyTorch package and building from the original work by \cite{williams2022data}. The source code and the notebooks used to produce the results of this paper are openly available at \href{https://github.com/ERMETE-Lab/NuSHRED}{github.com/ERMETE-Lab/NuSHRED}. Both the LSTM and SDN networks that comprise the SHRED architecture consist of two hidden layers: each hidden layer of the LSTM has 64 neurons, whereas the two hidden layers of the SDN have, respectively, 350 and 400 neurons, for a total of less than $1000$ network parameters.

The original SHRED architecture has been designed for reconstruction, prediction, and forecasting tasks for a single parametric configuration \cite{kutz_shallow_2024,riva2024robuststateestimationpartial,williams2022data}; the same architecture extends easily to parametric datasets with minimal modifications \cite{riva2025efficientparametricstateestimation,shredrom}. These adjustments mainly involve compressing the entire parametric dataset using the SVD, as SHRED naturally supports the inclusion of multiple trajectories corresponding to varying parameters; the LSTM then learns the proper dependence. In fact, SHRED can map the signals of the parametric measures to the corresponding latent representation. For large parametric datasets, compression via SVD is a crucial step to achieve short training times. In practice, given a snapshot matrix $\mathbb{X}^{\boldsymbol{\mu}_p}\in\mathbb{R}^{\mathcal{N}_h\times N_t}$ for a specific parameter $\boldsymbol{\mu}_p$, where $\mathcal{N}_h$ represents the spatial degrees of freedom and $N_t$ the number of time steps, the SVD computes a common basis $\mathbb{U}^{\boldsymbol{\mu}_p}\in\mathbb{R}^{\mathcal{N}_h\times r}$ of rank $r$. This basis allows the generation of a latent representation $\mathbb{V}^{\boldsymbol{\mu}_p}=\left(\mathbb{U}^{\boldsymbol{\mu}_p}\right)^T\mathbb{X}^{\boldsymbol{\mu}_p}\in\mathbb{R}^{r\times N_t}$, which embeds the (parametric) temporal dynamics required for SHRED training. When working with parametric datasets, a common basis capturing the dominant spatial physics is essential. If enough RAM is available, the snapshots for all parameters can be stacked to directly perform SVD onto the full dataset:
\begin{equation}
    \mathbb{X} = \left[\mathbb{X}^{\boldsymbol{\mu}_1}|\mathbb{X}^{\boldsymbol{\mu}_2}|\dots|\mathbb{X}^{\boldsymbol{\mu}_{N_p}}\right]
\end{equation}    

Otherwise, hierarchical SVD must be performed \cite{riva2025efficientparametricstateestimation}. For the present test case, the starting dataset comes from a one-dimensional code, and the spatial dimension is quite small. Despite the parametric nature of the dataset, the full snapshot matrix can be used to compute the SVD basis. The training and testing of the neural networks have been performed on a MacBook Pro (2024) with an M4 Pro chip installed. For completeness, both the compressive and full training approaches will be investigated and compared in the following sections.

% --------------------------------------
\subsection{Ensemble of SHRED models}
% --------------------------------------

The combination of compressive training and the small number of network parameters allows SHRED to be trained in minutes, even on personal laptops \cite{kutz_shallow_2024}. Then, it becomes computationally feasible to train different SHRED models for different sensor configurations, i.e., for different sensor triplets, to obtain more than one estimation of the state such that a measure of the uncertainty associated with the prediction can be provided \cite{riva2024robuststateestimationpartial}. This way of proceeding can be useful to generate state estimations which are robust against noisy measurements collected on the physical system itself; this operation mode is known as ensemble SHRED \cite{riva2024robuststateestimationpartial}. Let $\vec{Y}^{(k)}\in\mathbb{R}^{L\times s}$ be a set of $s$ temporal trajectories (with lag ${L}$) for the $k$-th configuration of sensors, such that the associated $k$-th SHRED model $\vec{f}_{(k)} = f_D^{(k)}(f_R^{(k)}(\cdot))$ can be trained to predict the state of the system:
\begin{equation}
    \vec{u} \approx \hat{\vec{u}}_{(k)} = \vec{f}_{(k)}(\vec{Y}^{(k)})
\end{equation}

By training $K$ different SHRED models on $K$ different sensor configurations, an ensemble of predictions to obtain the mean and covariance of the state estimation can be generated as follows:
\begin{subequations}
    \begin{align}
        \hat{\vec{u}} & = \frac{1}{K}\sum_{k=1}^K \hat{\vec{u}}_{(k)} \\
        \Sigma_{\vec{u}} & = \frac{1}{K-1}\sum_{k=1}^K (\hat{\vec{u}}_{(k)} - \hat{\vec{u}})^T(\hat{\vec{u}}_{(k)} - \hat{\vec{u}})
    \end{align}
\end{subequations}
The effect of the ensemble SHRED will be investigated in the following sections (for more details, the interested reader should refer to \cite{riva2024robuststateestimationpartial, riva2025efficientparametricstateestimation}).

% ---------------------------------
\section{DYNASTY Experimental Facility}\label{sec: dynasty}
% -------------------------------------

\begin{figure}
    \centering
    \includegraphics[width=1\linewidth]{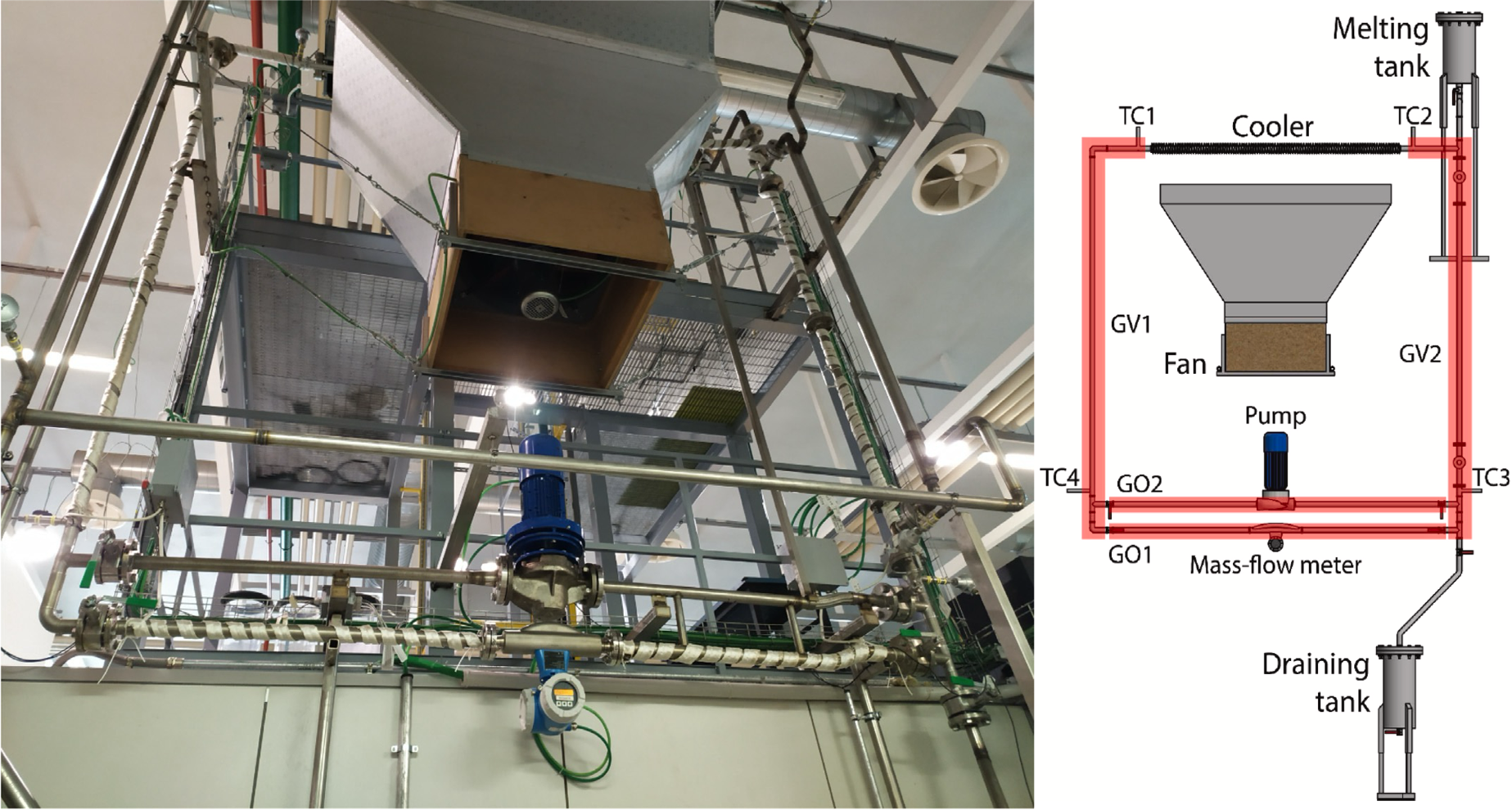}
    \caption{DYNASTY natural circulation loop. The photograph on the left shows the facility installed at the Energy Laboratories of Politecnico di Milano, while the schematic on the right highlights the loop layout and the main components, including the heated sections and the upper air-cooled heat sink.}
    \label{fig: dynasty-facility}
\end{figure}

DYNASTY (DYnamics of NAtural circulation for molten SalT internallY heated) is a natural circulation loop developed at the Energy Laboratories of Politecnico di Milano to investigate buoyancy-driven flows in the presence of a spatially distributed heat source \cite{benzoni2023preliminary}. The facility is primarily conceived as an experimental platform to reproduce, in a simplified and controlled manner, the thermal-hydraulic conditions relevant to internally heated systems, such as those encountered in molten salt reactor concepts.

The loop is realized entirely in AISI-316 stainless steel to ensure structural integrity at high temperatures and compatibility with various working fluids. All pipes have an internal diameter of 38 mm and a wall thickness of 2 mm. The overall geometry consists of four straight legs arranged in a rectangular configuration, as illustrated in Figure~\ref{fig: dynasty-facility}. Three of these legs are equipped with external electrical heating strips (GV1, GV2, and GO1-2), which are used to impose controlled heat inputs. Given the large ratio between the axial length and the pipe diameter, this heating strategy provides a suitable approximation of internal volumetric heat generation \cite{benzoni2023preliminary}. Heat removal is achieved through a finned horizontal tube located at the top of the loop, which acts as the system heat sink. The cooler can operate either in fully passive conditions or under forced convection by activating a fan installed below the finned section. This flexibility allows the facility to operate under different thermal boundary conditions, including both passive and actively enhanced cooling regimes.

By selectively powering the heating elements, DYNASTY operates under multiple experimental configurations. Localized heating scenarios can be imposed by activating a single leg, while distributed heating conditions are obtained by simultaneously activating all heated sections. In the present work, the RELAP5/MOD3.3 model of the DYNASTY facility is employed to simulate the four distinct heating configurations representative of the experimental campaign. These include two Vertical-Heater–Horizontal-Cooler (VHHC) cases, with either the left vertical leg (GV1) or the right vertical leg (GV2) acting as the heat source, one Horizontal-Heater–Horizontal-Cooler (HHHC) configuration with heating applied to the lower horizontal leg (GO1), and a Distributed Heating (DH) configuration in which all heated sections (GV1, GV2, and GO1) are simultaneously powered. In all cases, heat removal is provided by the upper finned pipe, with the cooling fan operating at a constant airflow rate of 3 m$^3$/s (corresponding to 75\% of the maximum fan speed).

The facility instrumentation includes four Type-J fluid thermocouples installed at the corners of the loop (TC1 to TC4, clockwise from the top left corner). These sensors provide local temperature measurements with an absolute accuracy of $\pm0.5\,^\circ\mathrm{C}$, to which electronic acquisition noise must be superimposed. The loop mass flow rate is measured in the center of the lower horizontal leg (GO1) using a Coriolis flow meter, characterized by an estimated uncertainty of $\pm0.25$~g/s. All signals are collected by a dedicated data acquisition and control system, which records measurements at a sampling rate of 1~Hz, applies basic signal conditioning (including spike removal), and stores the processed data in MATLAB-compatible formats .

% --------------------------------------
\subsection{The RELAP5 model of DYNASTY}
% --------------------------------------

\begin{figure}
    \centering
    \includegraphics[width=1\linewidth]{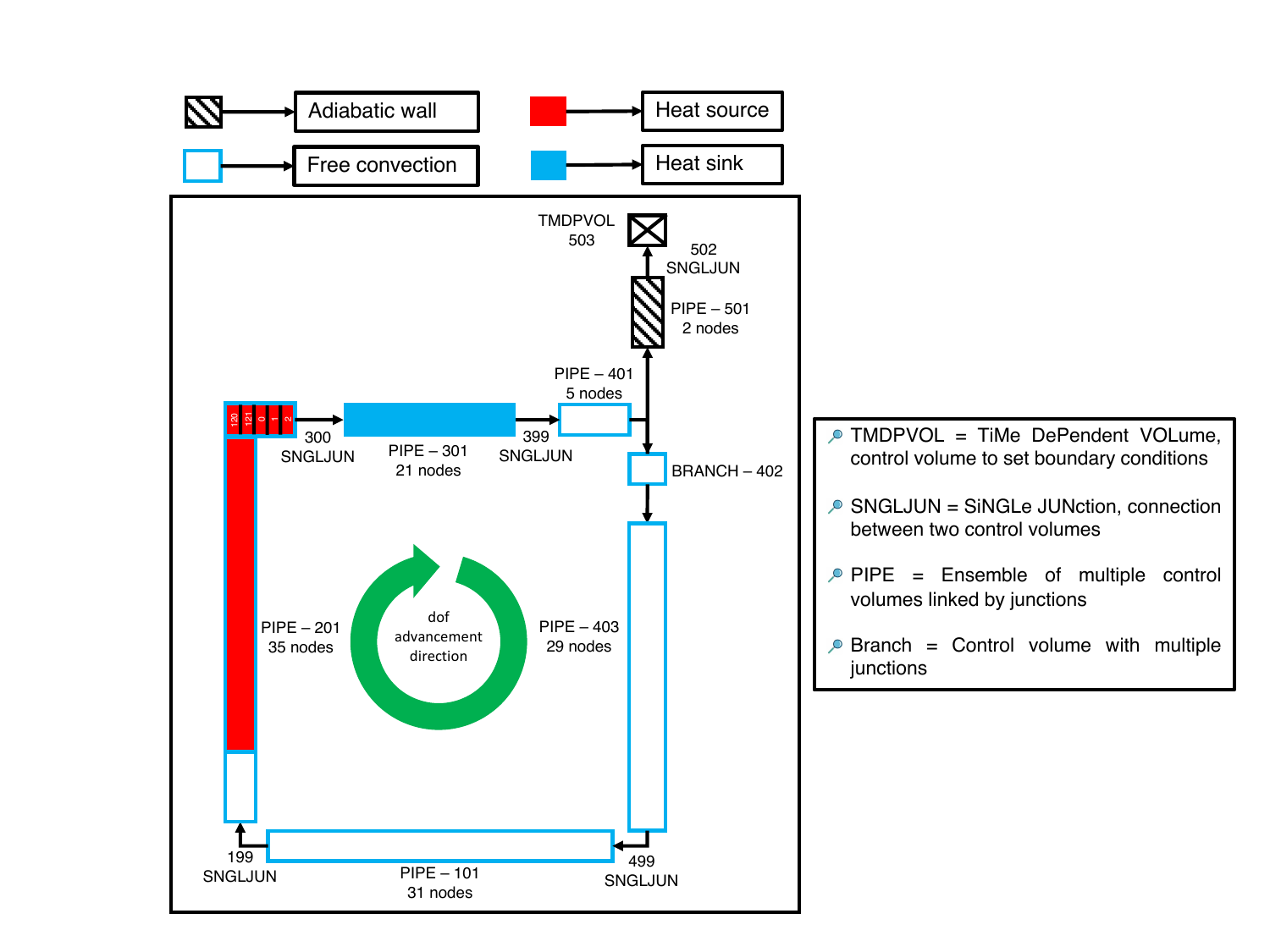}
    \caption{R5 nodalization of the DYNASTY facility used as a high-fidelity model for state estimation. The heated section corresponding to the VHHC-GV1 configuration is highlighted in red, while the upper air-cooled heat sink is shown in blue.}
    \label{fig: relap-nodes}
\end{figure}

The numerical framework adopted in this work is conceived to support the application of the Shallow Recurrent Decoder (SHRED) architecture to a real experimental facility. Within this context, the RELAP5/MOD3.3 system code is employed as a high-fidelity thermal–hydraulic model of the DYNASTY loop, providing physically consistent, space- and time-resolved data for state estimation. RELAP5 (R5) was originally developed at the Idaho National Engineering Laboratory for the U.S. Nuclear Regulatory Commission \cite{fletcher1992relap5} and is widely used for the simulation of transient thermal–hydraulic phenomena in nuclear and non-nuclear systems.

R5 is based on a one-dimensional, non-homogeneous, non-equilibrium two-fluid formulation, solved through a partially implicit numerical scheme. This modelling approach allows the code to capture the dominant physical mechanisms governing buoyancy-driven flows while remaining computationally efficient, a key requirement for generating large datasets suitable for reduced-order representations and data-driven architectures. The governing equations consist of separate conservation laws for mass, momentum, and energy for each phase, formulated over control volumes aligned along differential stream tubes. The dependent variables represent volume- and time-averaged quantities, while time and one spatial coordinate constitute the independent variables. A comprehensive description of the mathematical formulation implemented in R5 is reported in \cite{mangal2012capability}. 

The R5 nodalization of the DYNASTY facility adopted in this work is shown in Figure~\ref{fig: relap-nodes}. The experimental loop is represented by four PIPE components corresponding to the four legs of the facility. Dedicated heat structures are coupled to the hydrodynamic components to model both the heated legs and the upper finned cooler. Except for the cooler section, where heat removal is enhanced by forced convection due to the operation of the fan, all remaining pipes exchange heat with the surrounding environment through natural convection, modelled using the Churchill–Chu correlation \cite{bergman2011fundamentals}. A uniform axial discretization is adopted, with a control volume length of 100~mm, resulting in a total of 122 computational nodes. Sensitivity analyses on numerical methods and nodalization were performed in \cite{MISSAGLIA2025107466}.

The heating simulations start by applying the experimental heating power to the sections corresponding to the selected configuration, while simultaneously operating the cooling fan. This procedure reproduces the startup conditions observed during the experimental campaign \cite{benzoni2023preliminary}. The applied thermal power generates sufficient buoyancy forces to overcome both gravitational and frictional pressure drops along the loop. Distributed pressure losses are modelled by assuming a uniform pipe roughness of 50~$\mu$m, while localized pressure drops associated with elbows and the T-junction in the upper right corner of the loop are evaluated using standard correlations \cite{idelchik1987handbook}. The concentrated pressure loss introduced by the Coriolis mass flow meter (Endress+Hauser\textsuperscript{\tiny\textregistered} Promass F80 DN25) is implemented based on the provided technical data. The resulting R5 simulations well reproduce the transient establishment of natural circulation and provide full-field thermal-hydraulic data, which are subsequently used to train the SHRED architecture. 

\section{Numerical Results}\label{sec: num-res}

\begin{figure}
    \centering
    \includegraphics[width=1\linewidth]{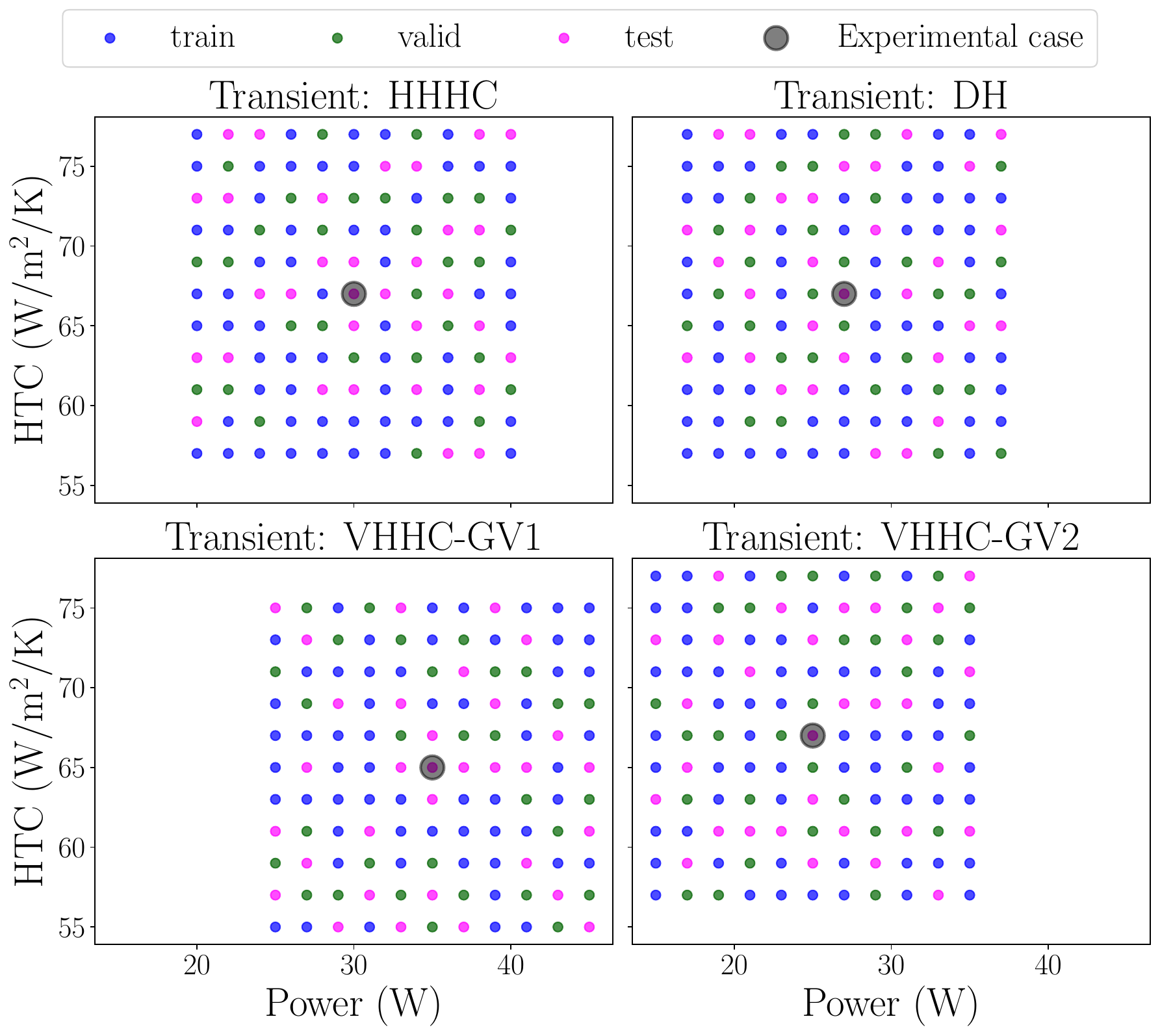}
    \caption{Division of the parametric space for training, validation and testing: the experimental case is always in the test set and in the middle of the parametric range.}
    \label{fig: parametric-space}
\end{figure}

The R5 simulations provide a rather large dataset of high-fidelity data, composed of the spatial temperature profile (stored as a vector of 122 elements) and the mass flow rate time series (a single value is sufficient due to the incompressibility of the flow and mass conservation). Hence, the state vector will have dimension $\mathcal{N}_h=123$. Four different experimental scenarios are simulated, corresponding to the four heating configurations described in Section \ref{sec: dynasty}, i.e., HHHC, two VHHC, and DH. For each one of these scenarios, experimental data of the temperatures at TC1-TC4 and the mass flow rate are available for specific values of the heat transfer coefficient (HTC) and the power provided to each control volume: for each scenario, starting from this central point, a parametric space is defined by varying the HTC and the power in a range of $\pm 10$ of the experimental values. A total of 121 simulations per scenario are performed, spanning 11 values for the HTC and 11 values for the power. The experimental case is always in the middle of the parametric range, as shown in Figure \ref{fig: parametric-space}. 

Figure \ref{fig: parametric-space} shows the division of the parametric space for training, validation, and test. Given a transient case $i$, the experimental point is always in the test set $\Xi^{\text{test}}_i$, whereas the training points, in $\Xi^{\text{train}}_i$ are randomly selected with ratio 50\%; the remaining $\sim$50\% of the points is further split into a validation set $\Xi^{\text{valid}}_i$ and test set $\Xi^{\text{test}}_i$, with a 50\%-50\% ratio. The network is optimized using the combination of the training points from all transients, as well as validated and tested 
\begin{equation}
\begin{split}
    \Xi^{\text{train}} &= \bigcup_{i}\Xi^{\text{train}}_i;\\
    \Xi^{\text{valid}} &= \bigcup_{i}\Xi^{\text{valid}}_i;\\
    \Xi^{\text{test}} &= \bigcup_{i}\Xi^{\text{test}}_i
\end{split} 
\end{equation}
In this way, the points are well distributed across the different transients. The validation set is used to select the best model during training, while the test set is used to evaluate the performance of the selected model on unseen model data. \\
Overall, three tensors of size $N_p \times N_t \times \mathcal{N}_h$ are generated, corresponding to the training, validation, and test sets, where $N_p$ is the number of parametric points in each set. The time window considered for each simulation is 3600 seconds, with a time step of 10 seconds, resulting in $N_t=361$ time steps for each trajectory (considering the initial condition as well). The data are normalized before training by applying a min-max normalization to each variable across the whole dataset. 

\begin{figure}
    \centering
    \includegraphics[width=1\linewidth]{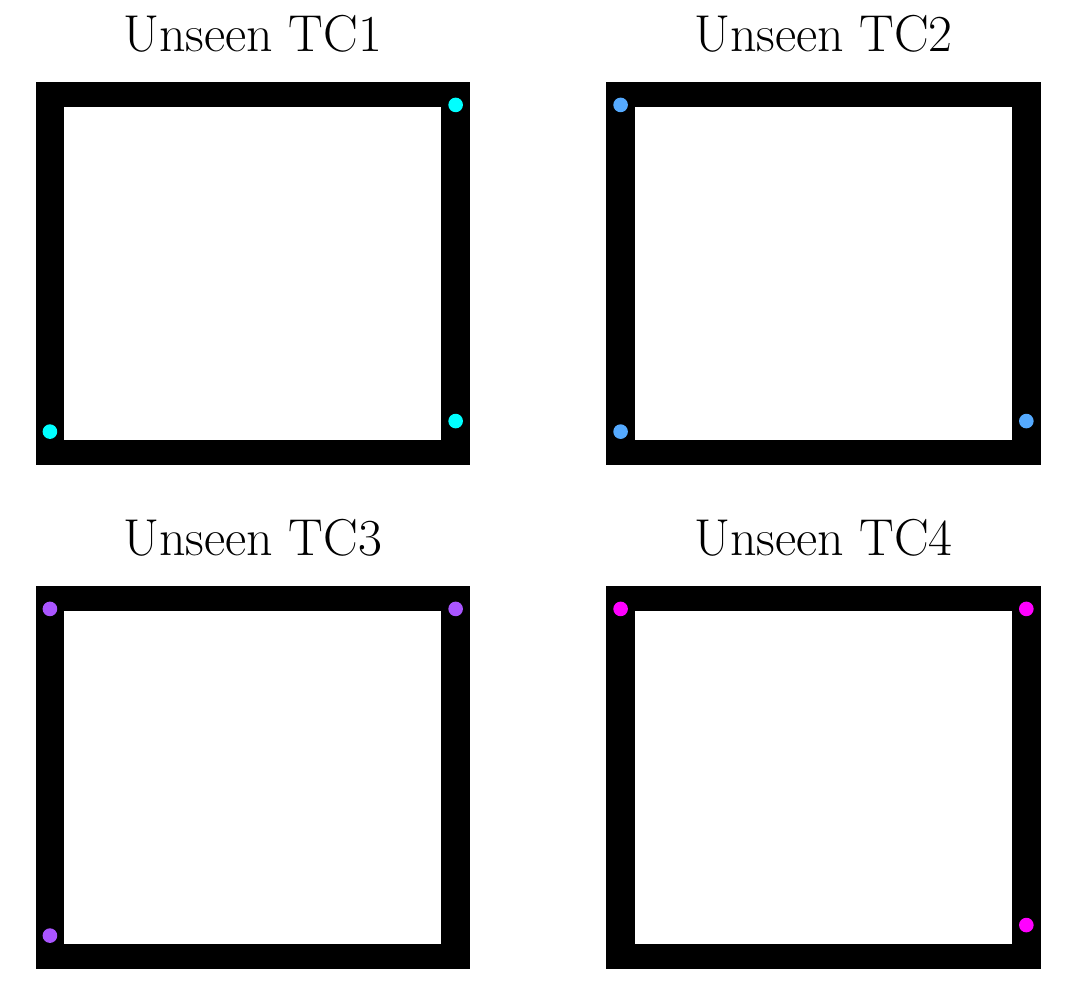}
    \caption{Definition of the 4 sensor configurations investigated in this work: in each one, three thermocouples are used as input to reconstruct the full state and one is left out for validation, in addition to the mass flow rate meter (not displayed).}
    \label{fig: sensors}
\end{figure}

For what concerns the input of the SHRED models, this work adopts a subset of three out of four thermocouples available. Figure \ref{fig: sensors} shows the four sensor configurations investigated in this work. In each one, three thermocouples are used to reconstruct the full state, and one is left out for validation, alongside the mass flow rate meter. The effect of the ensemble procedure will be discussed by ensembling the four SHRED models trained on the four possible sensor configurations. All SHRED networks have been trained on a Macbook Pro with Apple M4 Pro chip, using \textit{mps} (Metal Performance Shaders backend for GPU training acceleration).

In this work, two main analyses will be carried out: a parametric verification of the SHRED architecture with the R5 data, and a validation of the architecture on the experimental data. Moreover, in the latter, the network is tested at the experimental point in prediction regime for time instants beyond the training interval.

\subsection{Dimensionality Reduction and Training Times}

\begin{figure*}
    \centering
    \includegraphics[width=1\linewidth]{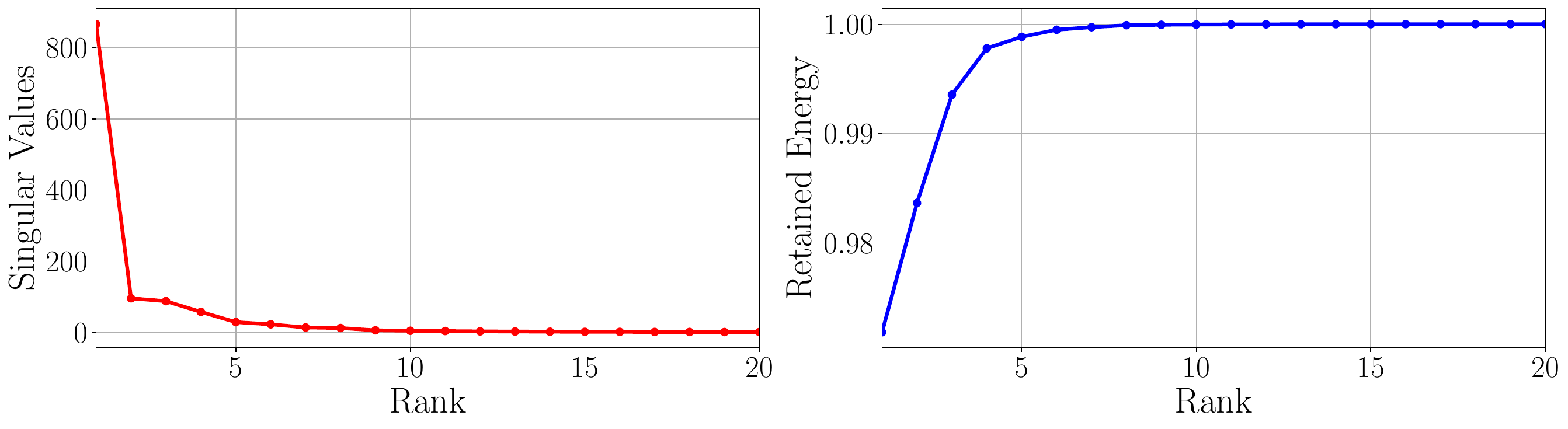}
    \caption{Decay of the singular values for the temperature field (left) and retained energy based on their cumulative sum (right).}
    \label{fig: svd-analysis}
\end{figure*}

As described in Section \ref{sec: shred}, the SHRED architecture can be trained either on the full state or on a compressed representation of the state obtained through the SVD of the snapshot matrix. Since $\mathcal{N}_h$ is not large, the full training approach is computationally affordable; for completeness, the compressive training will be investigated as well. Due to the rather small size of the stacked snapshot matrix of size $\mathcal{N}_h \times (N_t \cdot N_p^{\text{train}})$, the SVD is performed directly on the full snapshot matrix, without the need for a hierarchical approach. The decay of the singular values is shown in Figure \ref{fig: svd-analysis} (left), where it can be seen that the first 6 singular values are sufficient to retain more than 99\% of the energy of the system, as shown in Figure \ref{fig: svd-analysis} (right). Hence, a rank $r=6$ is selected for the compressive training.

\begin{table*}[t]
\centering
\begin{tabular}{llccc}
\toprule
Method & Configuration & Time (min) & Epochs & Time/Epoch (min) \\
\midrule
Full        & Unseen TC1 & 8.4324  & 76  & 0.1110 \\
Full        & Unseen TC2 & 7.5419  & 70  & 0.1077 \\
Full        & Unseen TC3 & 8.4453  & 78  & 0.1083 \\
Full        & Unseen TC4 & 10.2979 & 95  & 0.1084 \\
\midrule
\textbf{Average} &  & 8.6794 & 79.75 & 0.1088 \\
\textbf{Std}     &  & 1.1589 & 10.72 & 0.0014 \\
\midrule
Compressive & Unseen TC1 & 4.3938  & 50  & 0.0879 \\
Compressive & Unseen TC2 & 6.1928  & 74  & 0.0837 \\
Compressive & Unseen TC3 & 6.1078  & 69  & 0.0885 \\
Compressive & Unseen TC4 & 6.7840  & 76  & 0.0893 \\
\midrule
\textbf{Average} &  & 5.8696 & 67.25 & 0.0873 \\
\textbf{Std}     &  & 1.0288 & 11.87 & 0.0025 \\
\bottomrule
\end{tabular}
\caption{Training performance for full and compressive strategies, for a single SHRED model.}
\label{tab:training_comparison}
\end{table*}

Table \ref{tab:training_comparison} shows the training performance for the full and compressive strategies. The training time is comparable between the two approaches, with the compressive training being slightly more expensive due to the fact that more epochs are required to reach convergence. However, the time per epoch is very similar between the two approaches, with a negligible difference due to the fact that the high-dimensional dimension $\mathcal{N}_h$ is not sensibly larger than the rank $r$ of the compressive representation. The compressive training approach would show more significant computational advantages for larger datasets, where the high-dimensional dimension $\mathcal{N}_h \sim 10^4$ or more and the rank $r$ of the compressive representation is typically in the order of tens.

\subsection{Parametric Verification with the Full Order Model}

\begin{figure*}
    \centering
    \includegraphics[width=1\linewidth]{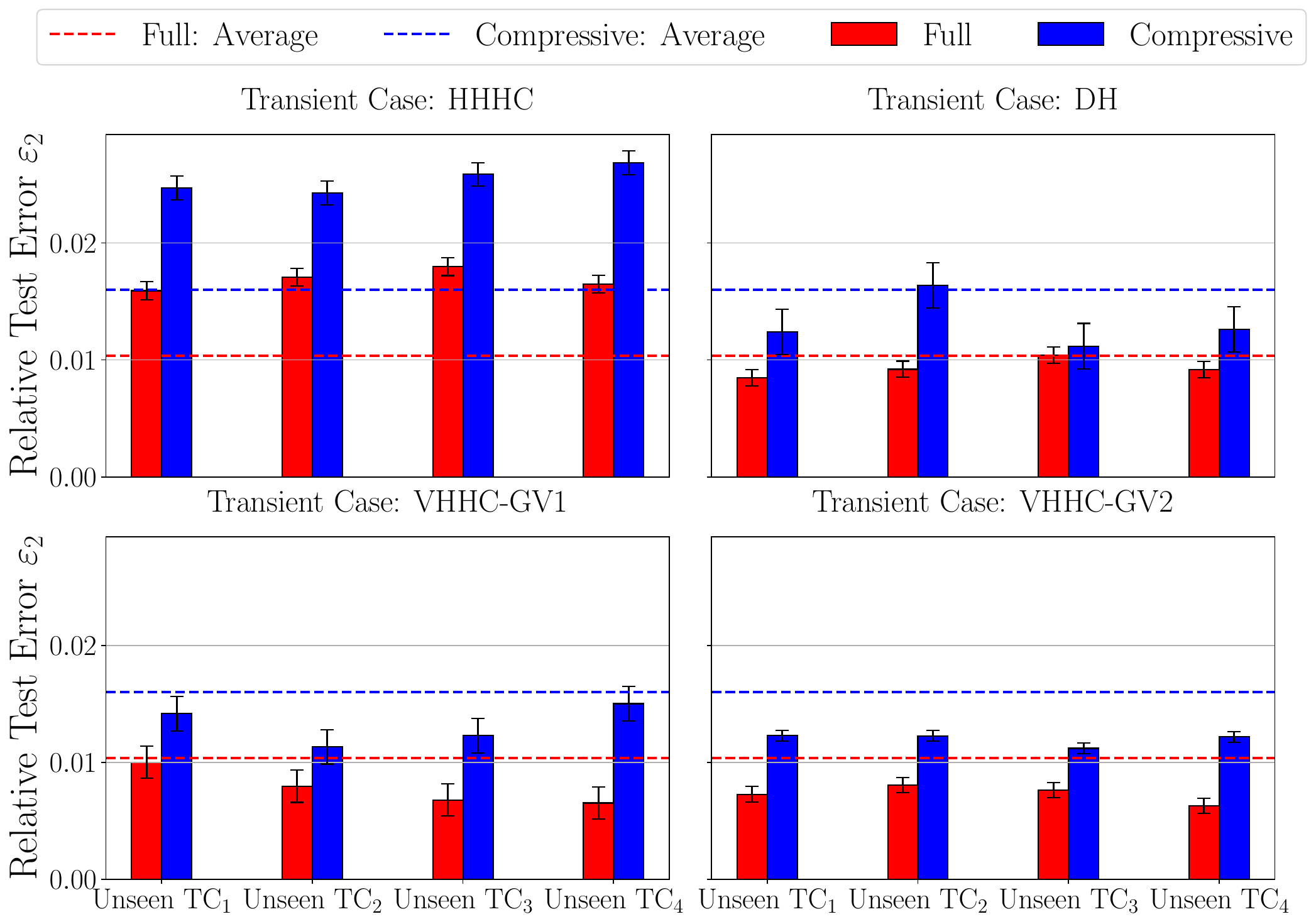}
    \caption{Relative errors of the SHRED reconstruction for the test set on each transient across the parametric space (bars), for both the full (red) and compressive (blue) training approaches. The dotted lines represents the average value of either the full or compressive approach over the different transients.}
    \label{fig: fom-rel-errs}
\end{figure*}

At first, the SHRED architecture is verified against simulation test data (see Figure \ref{fig: parametric-space}) by evaluating the relative error $\varepsilon_2$ of the reconstruction as follows:
\begin{equation}
    \varepsilon_2 = \frac{1}{\text{dim}(\Xi^{\text{test}})}\sum_{n\in\Xi^{\text{test}}}\frac{\norma{\vec{u}_n - \hat{\vec{u}}_n}_2}{\norma{\vec{u}_n}_2}
\end{equation}
where $\vec{u}_n$ is the true state of the system at the test index $n$, $\hat{\vec{u}}_n$ is the corresponding SHRED reconstruction, and $\Xi^{\text{test}}$ is the test set. The relative error is computed for each test point in the parametric space and then averaged across all test points to obtain a single measure of the reconstruction performance. The results are shown in Figure \ref{fig: fom-rel-errs}, where it can be seen that the relative errors are generally low, with values below 0.015 for all configurations and both training approaches. The compressive training approach shows slightly higher errors compared to the full training, which is expected due to the additional approximation introduced by the SVD compression. However, the errors remain within an acceptable range, demonstrating that the SHRED architecture is able to effectively capture the underlying dynamics of the system even when trained on a compressed representation of the state.

\begin{figure*}
    \centering
    \begin{overpic}[width=0.925\linewidth]{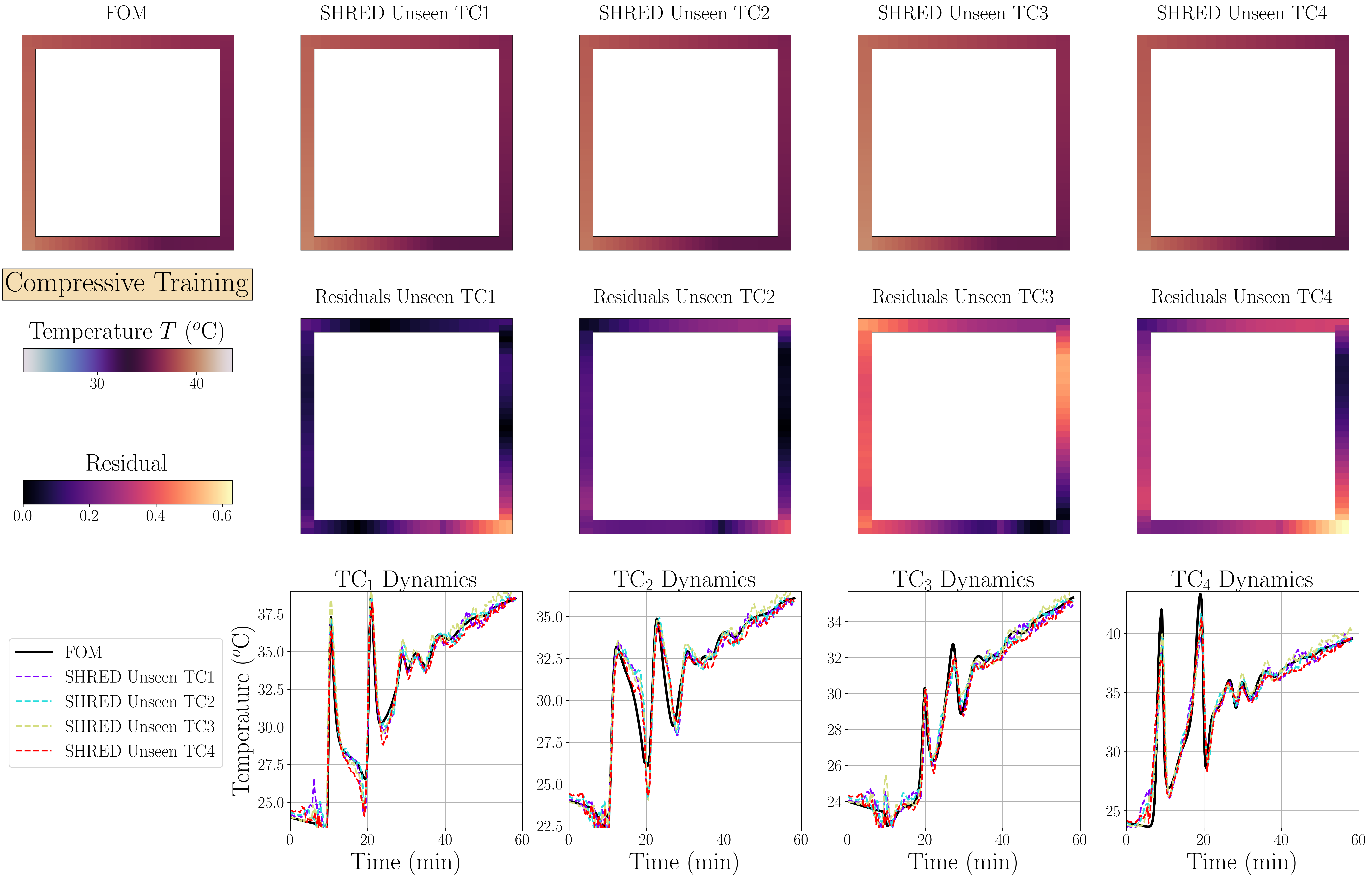}
    \put(3,3){(a)}
    \end{overpic}
    \begin{overpic}[width=0.925\linewidth]{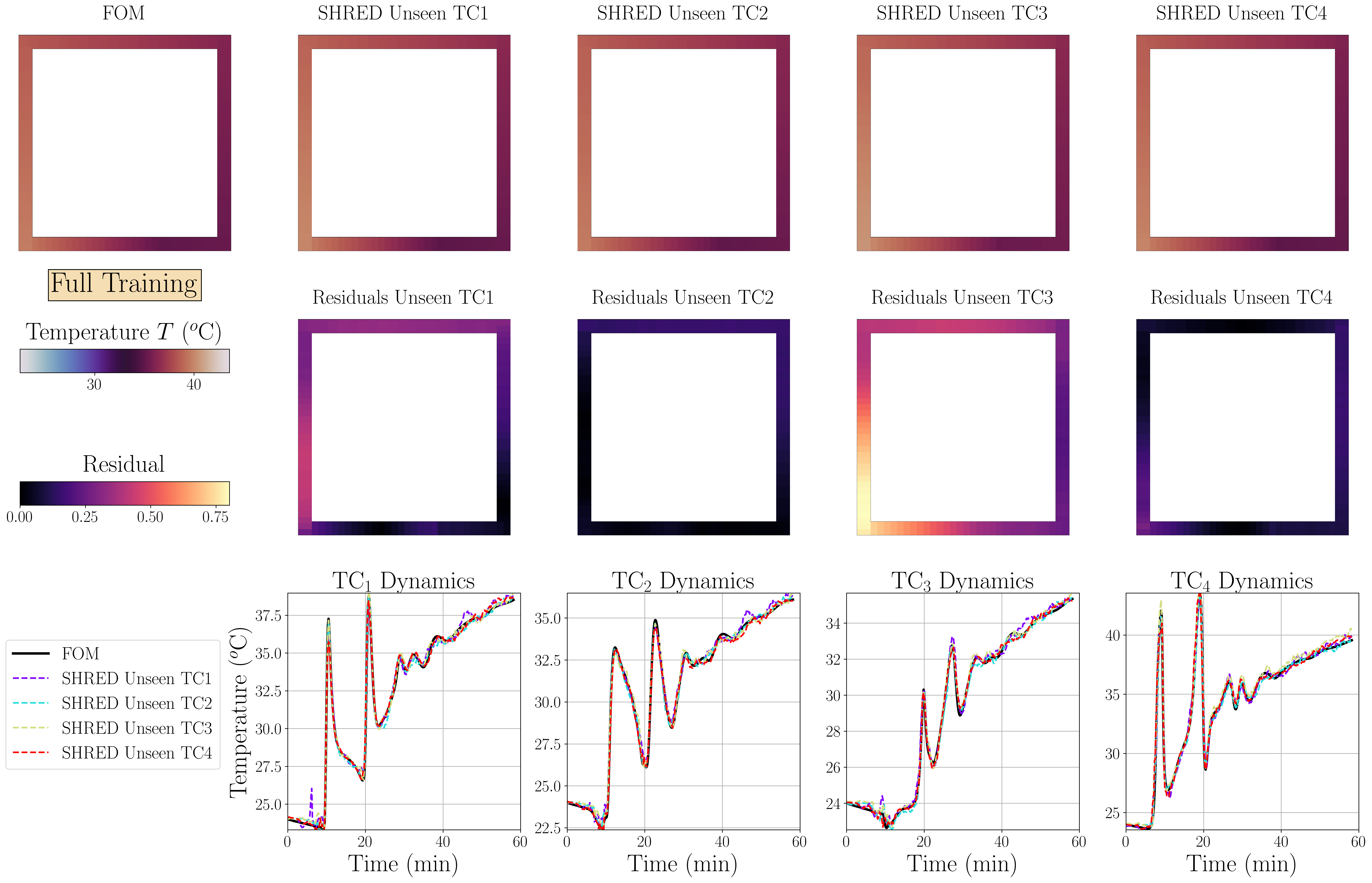}
    \put(3,3){(b)} 
    \end{overpic}
    \caption{Reconstruction of the temperature field at final time for the HHHC scenario at the experimental test point, for the compressive (a) and full (b) training approaches. The first row shows the contour plots of the true state (first column) and the SHRED reconstructions (from the second column onwards), the second row shows the corresponding error fields and the first column of the third row shows the temperature evolution over time for the true state and the SHRED reconstructions at the TC locations.}
    \label{fig: fom-reconstruction-shred}
\end{figure*}

Figure \ref{fig: fom-reconstruction-shred} shows the reconstruction of the temperature field at the final time for the HHHC scenario at the experimental test point, for both the compressive and full training approaches. This scenario has been chosen as the representative one due to the strong oscillations observed in the temperature field, which represent a challenging test case for the SHRED architecture; moreover, this case is the one with the highest errors (Figure \ref{fig: fom-rel-errs}). The contour plots of the true state and the SHRED reconstructions show a good agreement, for both the compressive and full training approaches, with a residual field of less than 0.75 degrees. By looking at the time evolution of the temperature at the thermocouples' locations, the SHRED reconstructions are able to capture the oscillatory behaviour of the system, with an almost perfect agreement on the unseen thermocouple: in particular, the last row shows the dynamics of the unseen thermocouple for the configuration which does not see it as input. This result is remarkable cause it shows that SHRED is able to discriminate between different scenarios and reconstruct the state properly. 

\begin{figure*}
    \centering
    \includegraphics[width=1\linewidth]{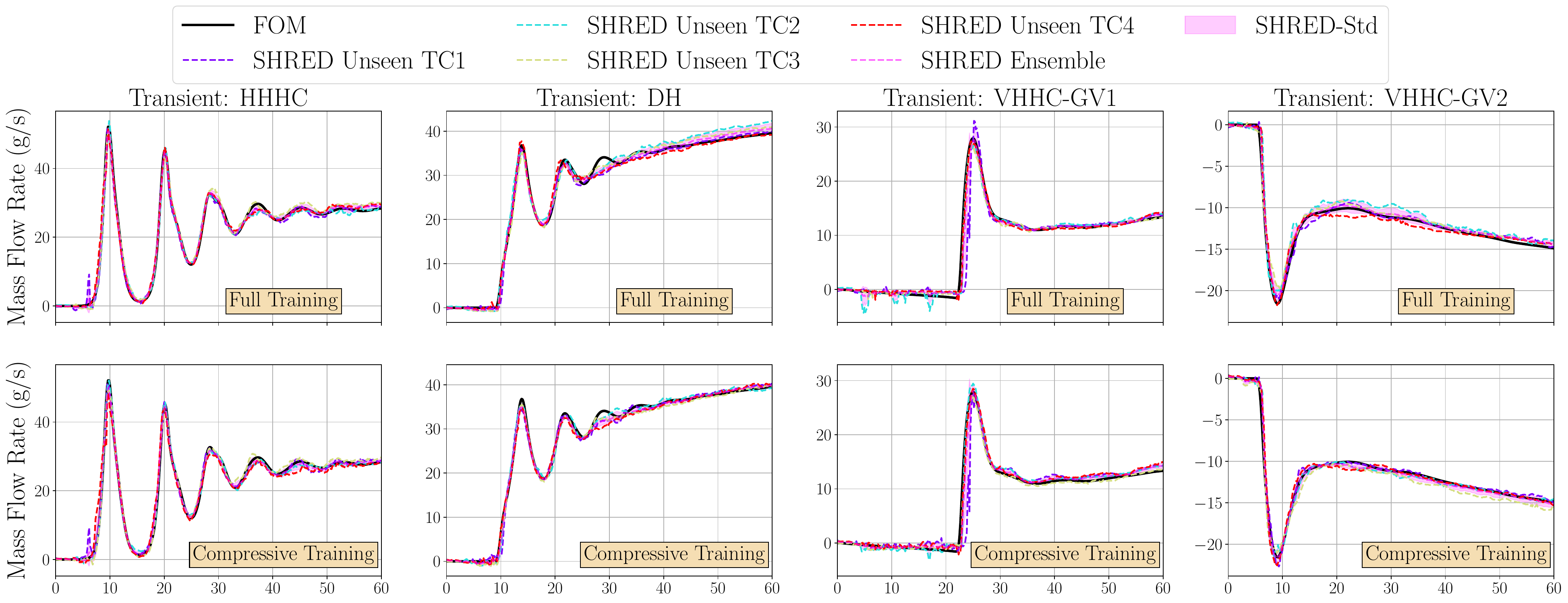}
    \caption{Comparison of the mass flow rate time series for all the scenarios at the experimental test point, between the R5 simulation and the SHRED reconstructions obtained with the compressive (first row) and full (second row) training approaches. This includes also the ensemble SHRED reconstruction.}
    \label{fig: fom-reconstruction-shred-mfr}
\end{figure*}

Furthermore, the comparison of the mass flow rate time series for scenarios at the experimental test point, shown in Figure \ref{fig: fom-reconstruction-shred-mfr}, confirms that SHRED is able to indirectly learn the map between the temperature field and the mass flow rate, even if only input with temperature measurements at the TC locations. This is a key result, as it demonstrates that SHRED can effectively leverage the information contained in the sensor measurements to reconstruct not only the temperature field but also other relevant variables of the system, such as the mass flow rate, which is not directly measured by the sensors. In the same Figure, the ensemble SHRED reconstruction is plotted as well, showing a smoother reconstruction of the mass flow rate time series, with a better agreement with the R5 simulation compared to the single SHRED reconstructions obtained through the compressive and full training approaches.

\subsection{Comparison with Experimental Data}

In this section, the SHRED architecture is validated against the experimental data collected at the DYNASTY facility. The R5 simulations predict the evolution for 1 hour, while experimental data are available for a longer time window, around 155 minutes. Hence, the SHRED reconstructions are extrapolated in time up to 155 minutes by feeding the LSTM with the real measurements collected at the thermocouples' locations for the whole time window, hence doing prediction. A first comparison between the SHRED reconstructions, the R5 FOM, and the experimental data is carried out by evaluating the relative RMSE (rRMSE) of the reconstruction. Let $y_i(t)$ be the true measurement at the $i$-th location at time $t$, $\hat{y}_i(t)$ be the corresponding SHRED reconstruction/FOM prediction, and $N_t$ be the number of time steps; the relative RMSE for sensor $i$ is then computed as follows:
\begin{equation}
    \text{rRMSE}_i = \sqrt{\frac{1}{N_t}\sum_{t=1}^{N_t} (y_i(t) - \hat{y}_i(t))^2}\cdot \frac{1}{\frac{1}{N_t}\sum_{t=1}^{N_t} |y_i(t)|}
\end{equation}

\begin{table*}[t]
    \centering
    \begin{tabular}{l||cccc|cc}
    \hline
    \textbf{Configuration} & \textbf{HHHC} & \textbf{DH} & \textbf{VHHC-GV1} & \textbf{VHHC-GV2} & \textbf{Avg} & \textbf{Std} \\
    \hline
    FOM                       & 0.127 & 0.143 & 0.069 & 0.053 & 0.098 & 0.044 \\ \hline
    
    Unseen TC1 -- Full        & 0.056 & 0.057 & 0.120 & 0.099 & 0.083 & 0.032 \\
    Unseen TC1 -- Comp.       & 0.057 & 0.058 & 0.115 & 0.098 & 0.082 & 0.029 \\ \hline
    
    Unseen TC2 -- Full        & 0.038 & 0.070 & 0.054 & 0.125 & 0.072 & 0.038 \\
    Unseen TC2 -- Comp.       & 0.045 & 0.087 & 0.120 & 0.128 & 0.095 & 0.038 \\ \hline
    
    Unseen TC3 -- Full        & 0.061 & 0.057 & 0.094 & 0.062 & 0.069 & 0.017 \\
    Unseen TC3 -- Comp.       & 0.051 & 0.064 & 0.061 & 0.058 & 0.058 & 0.005 \\ \hline
    
    Unseen TC4 -- Full        & 0.060 & 0.067 & 0.120 & 0.064 & 0.078 & 0.028 \\
    Unseen TC4 -- Comp.       & 0.065 & 0.065 & 0.163 & 0.089 & 0.095 & 0.046 \\ \hline
    
    Ensemble -- Full          & 0.046 & 0.055 & 0.063 & 0.076 & 0.060 & 0.013 \\
    Ensemble -- Comp.         & 0.046 & 0.054 & 0.051 & 0.080 & 0.058 & 0.015 \\
    \hline
    \end{tabular}
    \caption{Relative RMSE of the FOM and SHRED reconstructions for the experimental test point, for both the full and comp. training approaches. The ensemble SHRED reconstruction is also included. The last two columns show the average and standard deviation across sensors.}
    \label{tab:rmse_unseen}
\end{table*}
%    \label{tab:rmse_unseen}

Table \ref{tab:rmse_unseen} reports the average rRMSE, i.e. rRMSE $= \frac{1}{s}\sum_{i=1}^s \text{rRMSE}_i$, for the FOM and the SHRED reconstructions for the experimental scenarios, for both the full and compressive training approaches. The ensemble SHRED reconstruction is also included. The FOM predictions are not extrapolated in time, but they are directly compared with the experimental data for the simulation time window only, while the SHRED reconstructions are predicted in time up to 155 minutes. The ensemble mode outperforms the single SHRED reconstructions, with a lower average rRMSE. For the single SHRED reconstructions, the full training approach shows better performance compared to the compressive training, even though the differences between the two approaches are not very significant, and both approaches show a good agreement with the experimental data, with rRMSE values below 0.1 for all configurations and both training approaches. The FOM predictions show higher rRMSE values compared to the SHRED reconstructions, which is expected due to the fact that the FOM is not calibrated on the experimental data, and it may not capture all the complexities of the real system. Overall, these results demonstrate that the SHRED architecture can effectively leverage the information contained in the sensor measurements to reconstruct the state of the system and provide accurate predictions even when extrapolated in time.

\begin{figure*}
    \centering
    \begin{overpic}[width=0.7\linewidth]{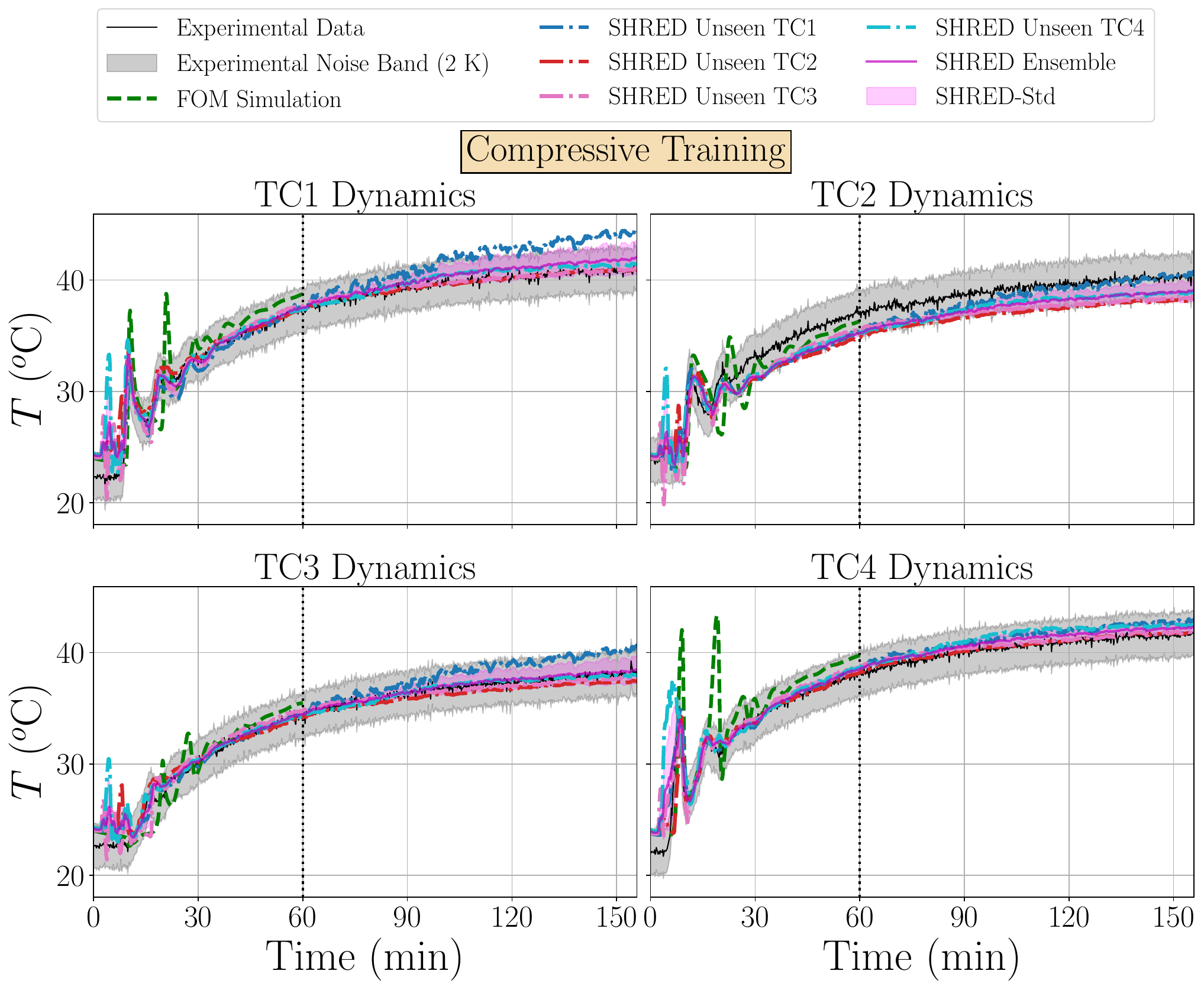}
    \put(7,67){(a)}
    \end{overpic}
    \begin{overpic}[width=0.7\linewidth]{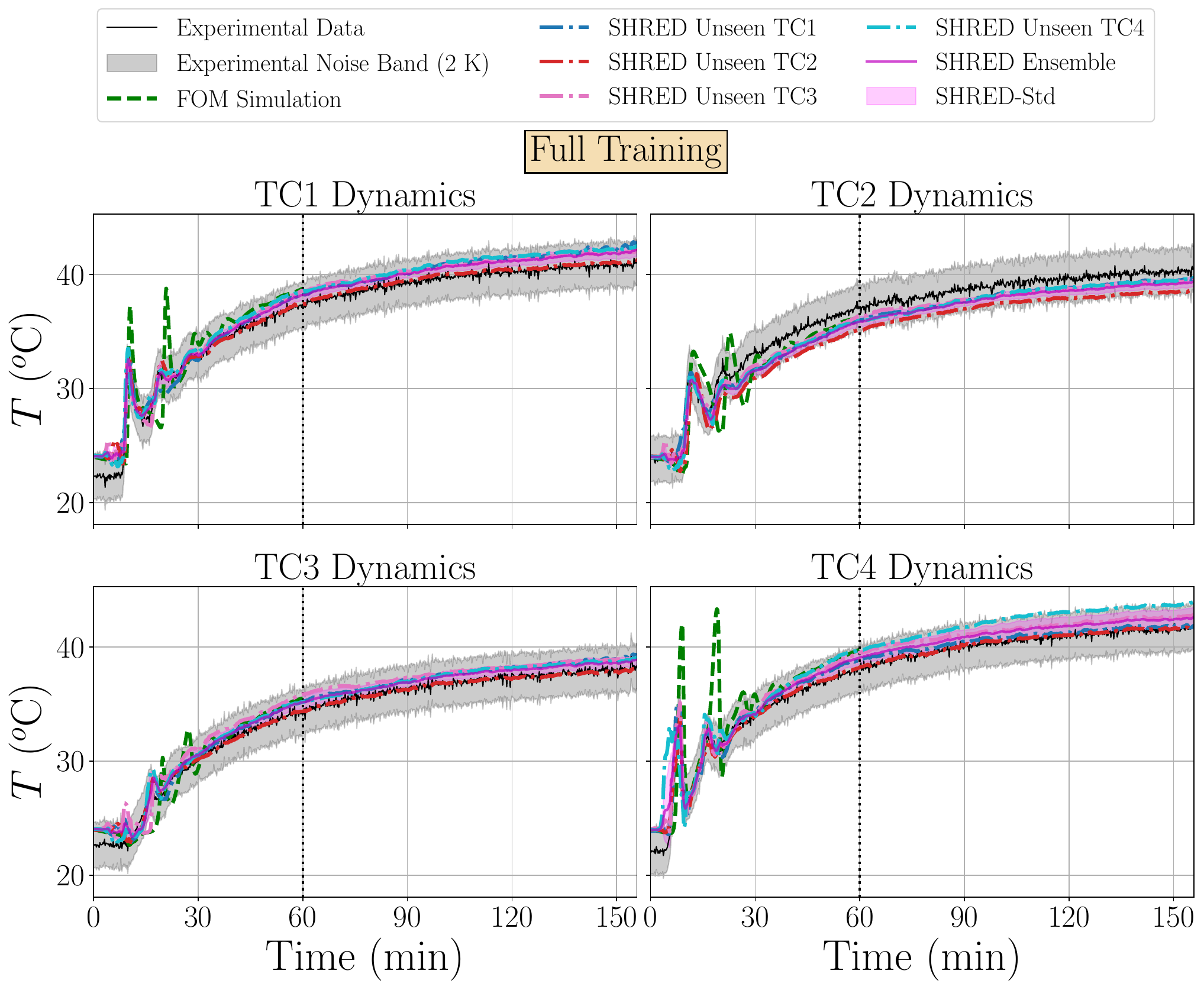}
    \put(7,67){(b)} 
    \end{overpic}
    \caption{Comparison of the TC measurements for the HHHC scenario at the experimental test point, between the experimental data, the FOM and the SHRED reconstructions obtained with the compressive (a) and full (b) training approaches. This includes also the ensemble SHRED reconstruction.}
    \label{fig: exp-reconstruction-shred-go1}
\end{figure*}

\begin{figure*}
    \centering
    \begin{overpic}[width=0.7\linewidth]{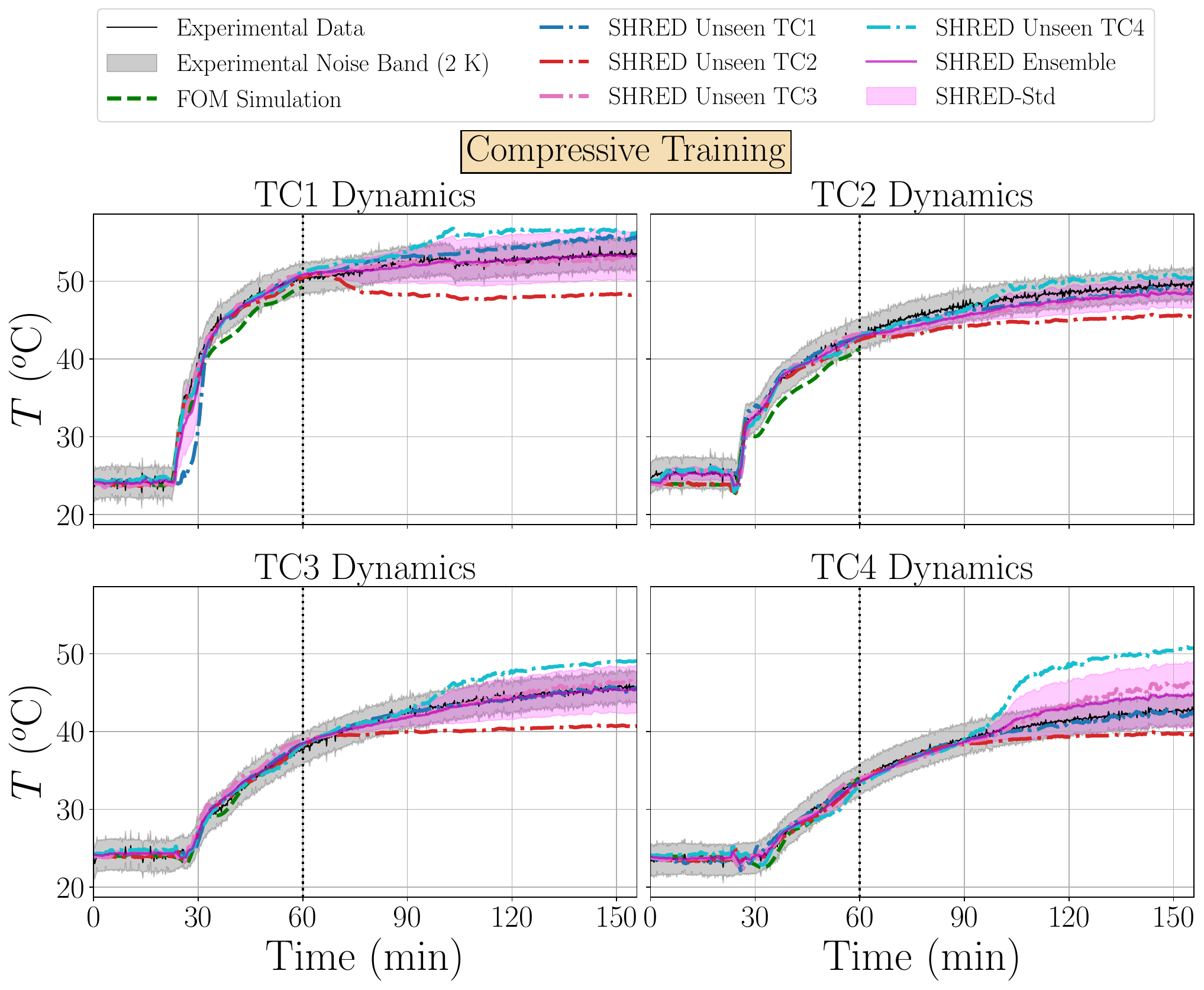}
    \put(7,67){(a)}
    \end{overpic}
    \begin{overpic}[width=0.7\linewidth]{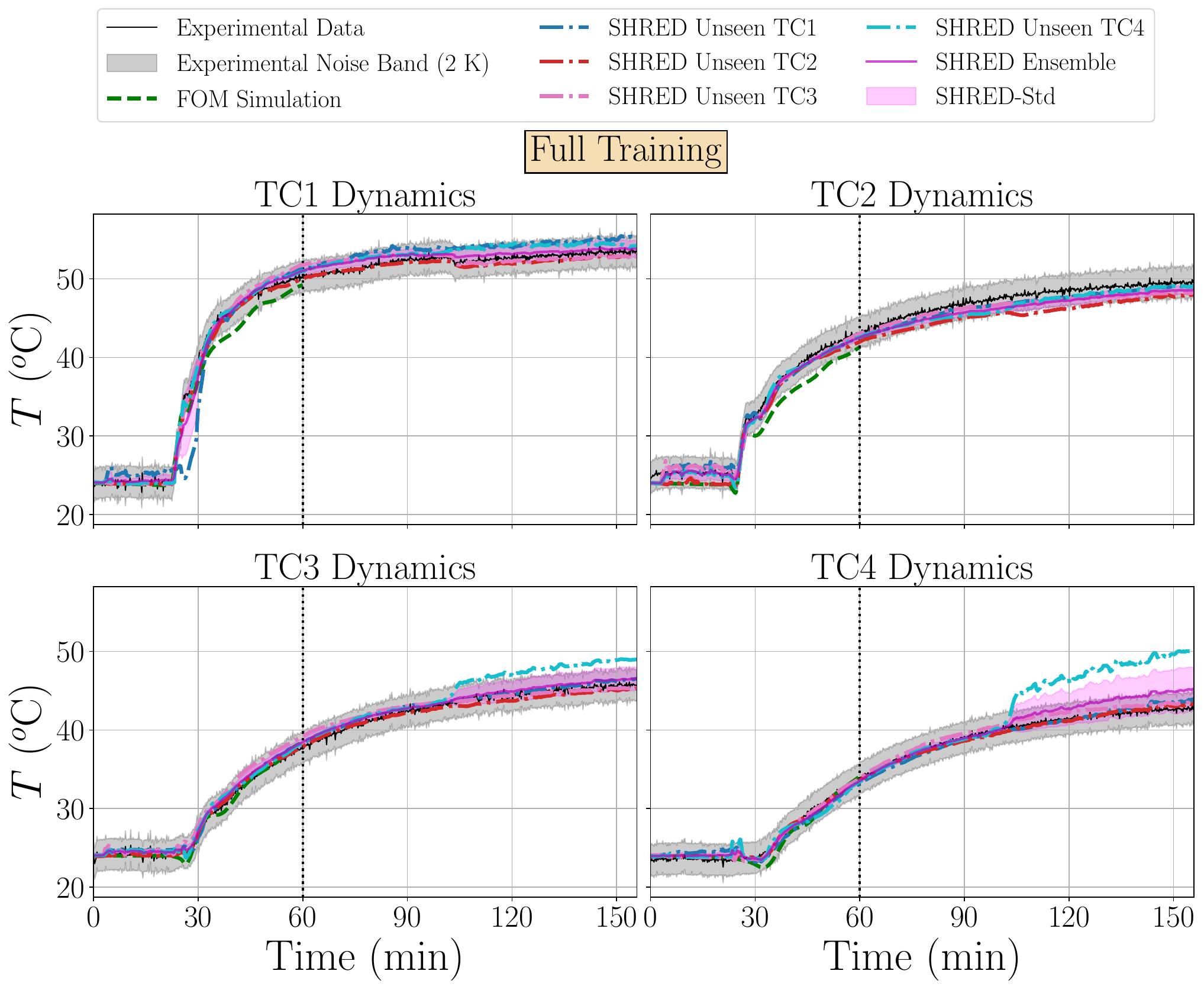}
    \put(7,67){(b)} 
    \end{overpic}
    \caption{Comparison of the TC measurements for the VHHC-GV1 scenario at the experimental test point, between the experimental data, the FOM and the SHRED reconstructions obtained with the compressive (a) and full (b) training approaches. This includes also the ensemble SHRED reconstruction.}
    \label{fig: exp-reconstruction-shred-gv1}
\end{figure*}

\begin{figure*}
    \centering
    \includegraphics[width=0.95\linewidth]{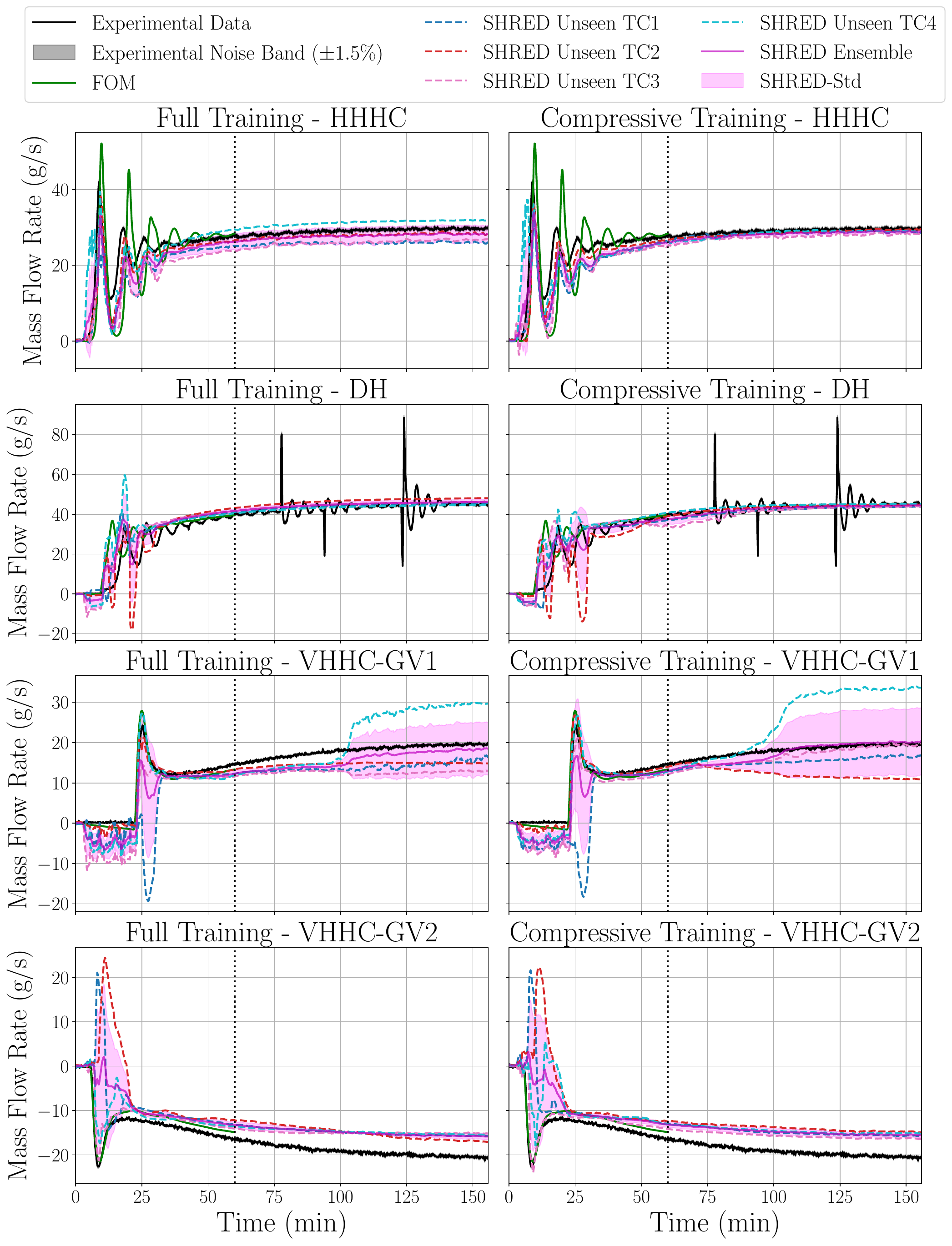}
    \caption{Comparison of the mass flow rate time series for all the scenarios at the experimental test point, between the experimental data, the FOM simulation and the SHRED reconstructions obtained with the full (first column) and compressive (second column) training approaches.}
    \label{fig: exp-reconstruction-shred-mfr}
\end{figure*}

Figures \ref{fig: exp-reconstruction-shred-go1} and \ref{fig: exp-reconstruction-shred-gv1} show the comparison of the TC measurements for the HHHC (the worst, with higher relative test error) and VHHC-GV1 (one of the others with similar errors) scenarios at the experimental test point, between the experimental data, the FOM, and the SHRED reconstructions obtained with both the compressive and full training approaches. The ensemble SHRED reconstruction is also included. For HHHC, the agreement between the SHRED reconstructions and the experimental data is very good, with the SHRED reconstructions being able to correctly capture the oscillations in the first hour, whereas the FOM overestimates the amplitude of the oscillations. A strong result is the fact that the SHRED reconstructions are able to extrapolate in time up to 155 minutes, by feeding the LSTM with the real measurements collected at the available thermocouple locations for the whole time window, and still maintain a good agreement with the experimental data. This shows how recurrent neural networks can effectively embed the dynamics of the measurements and produce a physical output, highlighting its natural capability of learning physics from data, even outside the training time interval, making these architecture very interesting for the development of monitoring pipelines within digital twins.

For VHHC-GV1, some discrepancies are observed at higher times between the SHRED reconstructions and the experimental data. Probably, this transient shows a more flat dynamics compared to the other cases, which is one of the drawbacks of LSTM units, since they are not able to discriminate between consecutive time instants. This can be directly measured by the standard deviation of the SHRED ensemble prediction: in fact, from 90 min onwards, the standard deviation increases significantly (especially for the compressive training approach), indicating a higher uncertainty in the SHRED reconstructions, passing from $\sim0.5$ K to $\sim2.5$ K. Overall, the performance of the SHRED reconstructions is still good, within the experimental uncertainty.

Finally, Figure \ref{fig: exp-reconstruction-shred-mfr} shows the comparison of the mass flow rate time series for all the scenarios at the experimental test point, between the experimental data, the FOM simulation, and the SHRED reconstructions. This is the most remarkable result, as it shows that SHRED is able to indirectly predict an observable quantity from the temperature measurements. As before, SHRED (both ensemble and single reconstructions) is able to go beyond in time and still maintain a good agreement with the experimental data. However, for the DH scenario, there are strong oscillations in the mass flow rate at $t=80$ and $t=120$ minutes, which are not captured by the SHRED reconstructions. This is due to the fact that the temperature does not oscillate as much as the mass flow rate, and therefore the LSTM is not aware of the oscillations in the mass flow rate, and therefore it cannot reconstruct them. Nevertheless, the average behaviour of the mass flow rate is well reconstructed, as seen by the RMSE values in Table \ref{tab:rmse_unseen}.

\section{Conclusions}\label{sec: concl}

This paper investigated the application of the Shallow Recurrent Decoder (SHRED) architecture for state estimation in a natural circulation loop, using both high-fidelity simulations and experimental data. The results demonstrate that SHRED is able to effectively leverage the information contained in the sensor measurements to reconstruct the state of the system and provide accurate predictions even when extrapolated in time, obtaining low reconstruction errors generally below 2\% with respect to the synthetic test dataset. The ensemble SHRED approach further improves the reconstruction performance by providing a measure of uncertainty associated with the predictions. Furthermore, the SHRED is a mature architecture able to be adopted in monitoring real systems, being trained with simulation data and tested on real experimental measurements SHRED is able to reconstruct seen and unseen sensor measurements with low RMSE $\sim0.1$. It is also capable of predicting the state of the system even beyond the training time interval. Overall, this work highlights the potential of SHRED as a powerful tool for state estimation in complex dynamical systems, with applications in various fields including thermal–hydraulic systems, fluid dynamics, and beyond. In the future, the SHRED architecture will be further investigated, focusing especially on its use for design applications and on the development of a data-assimilation framework based on SHRED for forecasting applications \cite{bao2025dataassimilationdiscrepancymodeling}.

%\clearpage

\section*{Code and supplementary materials}  
The code and data (compressed) are available at: \url{https://github.com/ERMETE-Lab/NuSHRED}.

% Bibliography
\bibliographystyle{unsrt}
\bibliography{bibliography}

\end{document}

%% file: z_vardef.tex
\DeclareMathAlphabet{\mathpzc}{OT1}{pzc}{m}{it}
 